\documentclass{article}

\usepackage{arxiv}
\usepackage[utf8]{inputenc} % allow utf-8 input
\usepackage[T1]{fontenc}    % use 8-bit T1 fonts
\usepackage{hyperref}       % hyperlinks
\usepackage{url}            % simple URL typesetting
\usepackage{graphicx}       % for including images
\usepackage{xcolor}         % colors
\usepackage{booktabs}       % professional-quality tables
\usepackage{amsfonts}       % blackboard math symbols
\usepackage{nicefrac}       % compact symbols for 1/2, etc.
\usepackage{microtype}      % microtypography
\usepackage{amsmath}        % for mathematical formatting
\usepackage{tabularray}     % for advanced tables
\SetTblrInner{rowsep=0pt}   % set table row spacing
\usepackage{bbm}            % blackboard-style bold math symbols
\usepackage{listings}       % for code formatting
\title{ReFINE: A Reward-Based Framework for Interpretable and Nuanced Evaluation of Radiology Report Generation}

\author{
Yunyi Liu \\
The University of Sydney \\
Sydney, NSW, Australia \\
\texttt{yunyi.liu1@sydney.edu.au} \\
\And
Yingshu Li \\
The University of Sydney \\
Sydney, NSW, Australia \\
\texttt{yingshu.li@sydney.edu.au} \\
\And
Zhanyu Wang \\
The University of Sydney \\
Sydney, NSW, Australia \\
\texttt{zhanyu.wang@sydney.edu.au} \\
\And
Xinyu Liang \\
Guangzhou University of Chinese Medicine \\
Guangzhou, China \\
\texttt{xinyu.liang31@gmail.com} \\
\And
Lingqiao Liu \\
The University of Adelaide \\
Adelaide, SA, Australia \\
\texttt{lingqiao.liu@adelaide.edu.au} \\
\And
Lei Wang \\
The University of Wollongong \\
Wollongong, NSW, Australia \\
\texttt{leiw@uow.edu.au} \\
\And
Luping Zhou \\
The University of Sydney \\
Sydney, NSW, Australia \\
\texttt{luping.zhou@sydney.edu.au} \\
}

\begin{document}
\maketitle
\begin{abstract}
Automated radiology report generation (R2Gen) has advanced significantly, introducing challenges in accurate evaluation due to its complexity. Traditional metrics often fall short by relying on rigid word-matching or focusing only on pathological entities, leading to inconsistencies with human assessments.
To bridge this gap, we introduce ReFINE, an automatic evaluation metric designed specifically for R2Gen. Our metric utilizes a reward model, guided by our margin-based reward enforcement loss, along with a tailored training data design that enables customization of evaluation criteria to suit user-defined needs. It not only scores reports according to user-specified criteria but also provides detailed sub-scores, enhancing interpretability and allowing users to adjust the criteria between different aspects of reports. Leveraging GPT-4, we designed an easy-to-use data generation pipeline, enabling us to produce extensive training data based on two distinct scoring systems, each containing reports of varying quality along with corresponding scores. These GPT-generated reports are then paired as accepted and rejected samples through our pairing rule to train an LLM towards our fine-grained reward model, which assigns higher rewards to the report with high quality. Our reward-control loss enables this model to simultaneously output multiple individual rewards corresponding to the number of evaluation criteria, with their summation as our final ReFINE. Our experiments demonstrate ReFINE's heightened correlation with human judgments and superior performance in model selection compared to traditional metrics. Notably, our model provides both an overall score and individual scores for each evaluation item, enhancing interpretability. We also demonstrate its flexible training across various evaluation systems. 
\end{abstract}

% keywords can be removed

\section{Introduction}
\label{sec:intro}
%Currently, automated radiology report generation (R2Gen) has experienced significant expansion~\cite{wang2023r2gengpt, li2024kargen}. This intricate AI task demands a profound comprehension of clinically relevant high-level semantics, presenting challenges in the generation process and evaluating the quality of the output reports. Automated assessment of radiology report generation typically involves metrics gauging the semantic accuracy of the generated reports against the reference reports. Traditional natural language generation (NLG) metrics,
%such as the BLEU metric~\cite{papineni2002bleu} and METEOR~\cite{reimers2019sentencebert}, primarily quantify n-gram matches, often overlooking important factors like lexical and structural diversity, which are essential for capturing the true meaning of the reports.
%These metrics are often criticized as misjudging paraphrasing and failing to capture complex diagnostic information adequately. 
Automated radiology report generation (R2Gen), which produces free-text descriptions about visual findings in radiographic images, has seen substantial growth~\cite{wang2023r2gengpt, li2024kargen}. This complex AI task requires understanding high-level clinical semantics, challenging both the generation and quality assessment of reports. R2Gen evaluation typically involves metrics assessing semantic accuracy of the generated reports against the ground-truth ones. Traditional natural language generation (NLG) metrics like BLEU~\cite{papineni2002bleu} and METEOR~\cite{reimers2019sentencebert} focus on n-gram matches, often missing crucial factors like lexical and structural diversity needed for true meaning. These metrics are often criticized for misjudging paraphrasing and failing to capture complex diagnostic details. To address these gaps, BERTScore~\cite{zhang2020bertscore} leverages contextualized embeddings to better capture paraphrasing, and clinical scores like CheXbert~\cite{smit2020chexbert} or Radgraph\_F1~\cite{jain2021radgraph} scores add clinically relevant dimensions by focusing on predefined pathological entities. 
%To address these issues, approaches like BERTScore~\cite{zhang2020bertscore} have been proposed, utilizing contextualized token embedding to detect paraphrasing more effectively. Furthermore, comprehensive evaluations now often incorporate clinically relevant scores, such as F1 scores of pathological entities labeled by CheXbert~\cite{smit2020chexbert} or Radgraph~\cite{jain2021radgraph}. However, these clinical scores are constrained by their predefined set of pathological entities and encounter challenges in accurately assessing the correlations among these entities. 
Despite efforts to improve the evaluation of report generation, existing evaluation metrics often do not align well with human judgment~\cite{liu2024systematic}. A recent work~\cite{yu2023evaluatingradcliq} proposed the RadCliQ score, which linearly combines multiple existing metrics while regressing combination weights from human-marked error scores to better align with human evaluation. However, RadCliQ's reliance on a limited set of expensive human-annotated training samples poses a challenge. While recent advances in Large Language Models (LLMs), like GPT-4~\cite{OpenAI2023GPT4TR}, suggest potential for report evaluation with proper prompts, directly applying GPT-4 for this purpose may be impractical. It raises privacy concerns due to online evaluation and demands substantial computing resources, considering its size and general-purpose nature, which may not be cost-effective for R2Gen.

To drive progress in this field, this study proposes ReFINE, an innovative metric tailored specifically for evaluating automated radiology report generation.
Leveraging GPT-4’s human-like scoring capacity~\cite{chiang2023can,liu2024mrscore}, our method autonomously produces evaluation samples that mimic human judgment. These samples are subsequently utilized to train an LLM-based reward model for automated scoring. In comparison to traditional evaluation metrics, ReFINE substantially improves the alignment with human assessments, leading to a more precise evaluation of report quality. Moreover, instead of merely providing an overall score, our model simultaneously outputs the scores for individual evaluation criteria, improving the interpretability of the assessment results. For example, by combining sub-criteria, we can pinpoint the reasons for a report’s poor quality, e.g., whether due to incorrect lesion location, incorrect severity of findings, or omission of findings. 
%Meanwhile, by generating training samples using LLMs, our method reduces the dependence on costly human annotations, enabling scalable model training and greater flexibility in adapting to different evaluation criteria. To operationalize our approach, we utilized two distinct sets of evaluation criteria (scoring systems) in this study.
Meanwhile, through LLM-generated samples, our method reduces dependence on costly human annotations, supporting scalable training and adaptability to various criteria. Utilizing the defined criteria, we prompt GPT-4 to generate report samples with varied quality levels, pairing reports of different quality corresponding to the same ground-truth report as "accepted" and "rejected" samples with score margins. These paired samples were then used to fine-tune the pretrained Llama3 model~\cite{meta2024llama} using reward modeling techniques. 
Our proposed loss function enables this model to produce multiple individual rewards concurrently, each corresponding to one evaluation criteria, which are then summed to produce our final ReFINE. Validating our model on two datasets paired with human evaluations, we found ReFINE aligns more closely with human judgment than other traditional metrics and exhibits versatility to accommodate different evaluation criteria.

Our ReFINE metric offers key advantages over existing approaches. Non-trainable metrics, including NLG-based and clinically relevant ones, correlate poorly with human assessments and cannot adapt to customized criteria. Among trainable metrics, RadCliQ combines non-trainable scores linearly, offering limited improvement and flexibility. Some LLM-based metrics (e.g., G\_Rad~\cite{chaves2024training} and FineRadScore~\cite{huang2024fineradscore}), relying on online LLMs, present privacy concerns. The most similar methods, MRScore~\cite{liu2024mrscore} and GREEN~\cite{ostmeier2024green}, fall short in different ways: MRScore only provides an overall score, while GREEN adds interpretability through free-text analysis but sacrifices scoring accuracy due to the task complexity. Furthermore, GREEN’s fine-tuning of LLMs lacks a dedicated loss function as ours, limiting its sensitivity to nuanced quality differences. Our method achieves a higher Kendall’s Tau correlation with human ratings (0.75 vs. 0.64 for GREEN), while reducing training costs (1x NVIDIA A6000 vs. 8x NVIDIA A100) and inference time. To date, no other trainable metrics demonstrate adaptability to different evaluation criteria as ours.

Our main contributions are summarized as follows:
\begin{itemize}
    \item[(1)] Our study presents a novel approach to training LLMs to generate ReFINE, a human-consistent metric designed for automated radiology report evaluation. Through our novel loss function discerning report rankings, we finetune LLMs to produce rewards aligned with our scoring system in a fine-grained manner, enhancing alignment with human evaluations and bolstering assessment accuracy. 
    \item[(2)] Importantly, our evaluation metrics assess not only the overall score for a report but also concurrently the detailed sub-scores based on diverse criteria. It enhances the interpretability of the evaluation, enabling users to pinpoint specific aspects influencing the overall score. 

    \item[(3)]  By facilitating a tailored analysis of report components, our ReFINE allows users to customize the evaluation framework to suit their specific needs. This level of customization could contribute to more targeted improvements in report generation. This capacity of ReFINE has been demonstrated by its versatility to accommodate two distinct sets of evaluation standards, respectively.

\end{itemize}

\section{Related Work}
\label{sec:relate}
\subsection{Evaluation Metrics for Radiology Reports}
Radiology report metrics can be categorized as language metrics and clinical metrics.

\noindent \textbf{Language Metrics.}~~for radiology report evaluations typically rely on structured assessments and direct comparison metrics. Common approaches like BLEU~\cite{papineni2002bleu}, ROUGE~\cite{lin2004rouge}, and METEOR~\cite{banerjee2005meteor} scores assess the textual similarity between the generated reports and a set of reference reports, focusing on aspects like n-gram overlap, precision, and recall. Other metrics like BERTScore~\cite{zhang2019bertscore} are calculated using embedding generated by pre-trained models to measure the similarity between the ground truth report and the generated report.  However, these methods have significant drawbacks. Firstly, they often do not capture the clinical relevance or the diagnostic accuracy of the content, as they primarily focus on linguistic features rather than medical correctness. 
Furthermore, when applied to evaluating text generated by large language models (LLMs), such as those based on GPT architectures, these traditional metrics fall short. The complexity and variability of text generated by LLMs mean that simple lexical or syntactic comparisons are insufficient. LLMs can generate clinically plausible text that may be lexically varied but semantically similar to the reference standards. This variability can lead to evaluations that are not reflective of actual clinical usability or accuracy. 

\noindent \textbf{Clinical Metrics.}~~focus more on the clinical description in the radiology report. One prevalent metric in contemporary research is CheXpert~\cite{irvin2019chexpert}, which mandates the extraction and labeling of 14 pathological entities as `present,' `absent,' or `uncertain.' The accuracy of these labels is typically assessed using tools like CheXbert, which also utilizes cosine similarity from embeddings as a metric. Another common method is RadGraph~\cite{jain2021radgraph}, which identifies clinical entities and their relationships within reports. However, these extraction-based techniques are constrained by a fixed set of entities and strict matching rules, which can lead to issues with coverage and difficulty addressing the ambiguous cases often found in reports. Although some hybrid approaches, such as RadCliQ and RadEval, attempt to amalgamate various metrics, they too fall short of fully capturing the nuances of clinical descriptions due to the inherent limitations of extraction-based methods.

\subsection{Large Language Model for Evaluation}
Previous methods such as G\_Rad~\cite{chaves2024training} and FineRadScore~\cite{huang2024fineradscore} leverage online LLMs like GPT-4 and Claude-3 for radiology report evaluation, achieving strong Kendall’s Tau correlations. However, these methods raise privacy concerns and rely on online access, limiting their practical application. On the other hand, GREEN~\cite{ostmeier2024green}, a parallel work to ours, addresses these concerns by using an offline model, as we do. Yet, GREEN generates free-text explanations with counts of clinical errors and matched findings, adding explainability but at the cost of complexity, which hampers its score prediction accuracy. Moreover, while GREEN fine-tunes pretrained LLMs, it lacks a dedicated loss function designed to capture subtle quality differences, as seen in our method. Consequently, our approach not only outperforms GREEN with a higher Kendall’s Tau correlation with human ratings (0.75 vs. 0.64) but also requires significantly lower computational resources (1x NVIDIA A6000 vs 8x NVIDIA A100 for training) and inference time. By incorporating multiple evaluation criteria and breaking down scores into granular components, our model enhances interpretability, allowing users to identify which specific reason contributes to the overall score, with a high alignment to human assessments.
\section{Method}\label{sec:method}
Traditional NLP evaluation metrics typically assess the similarity between a machine-generated report $x$ and a reference report $\hat{x}$ using n-gram overlap. However, these metrics often fail to capture the semantic equivalence and clinical relevance essential for accurate radiology report evaluation. To address these shortcomings, we introduce a new evaluation metric that better reflects the semantic content and clinical significance of the reports, aligning closely with human assessments. 

\begin{figure*}[t]
\centering
\centerline{\includegraphics[width=\linewidth]{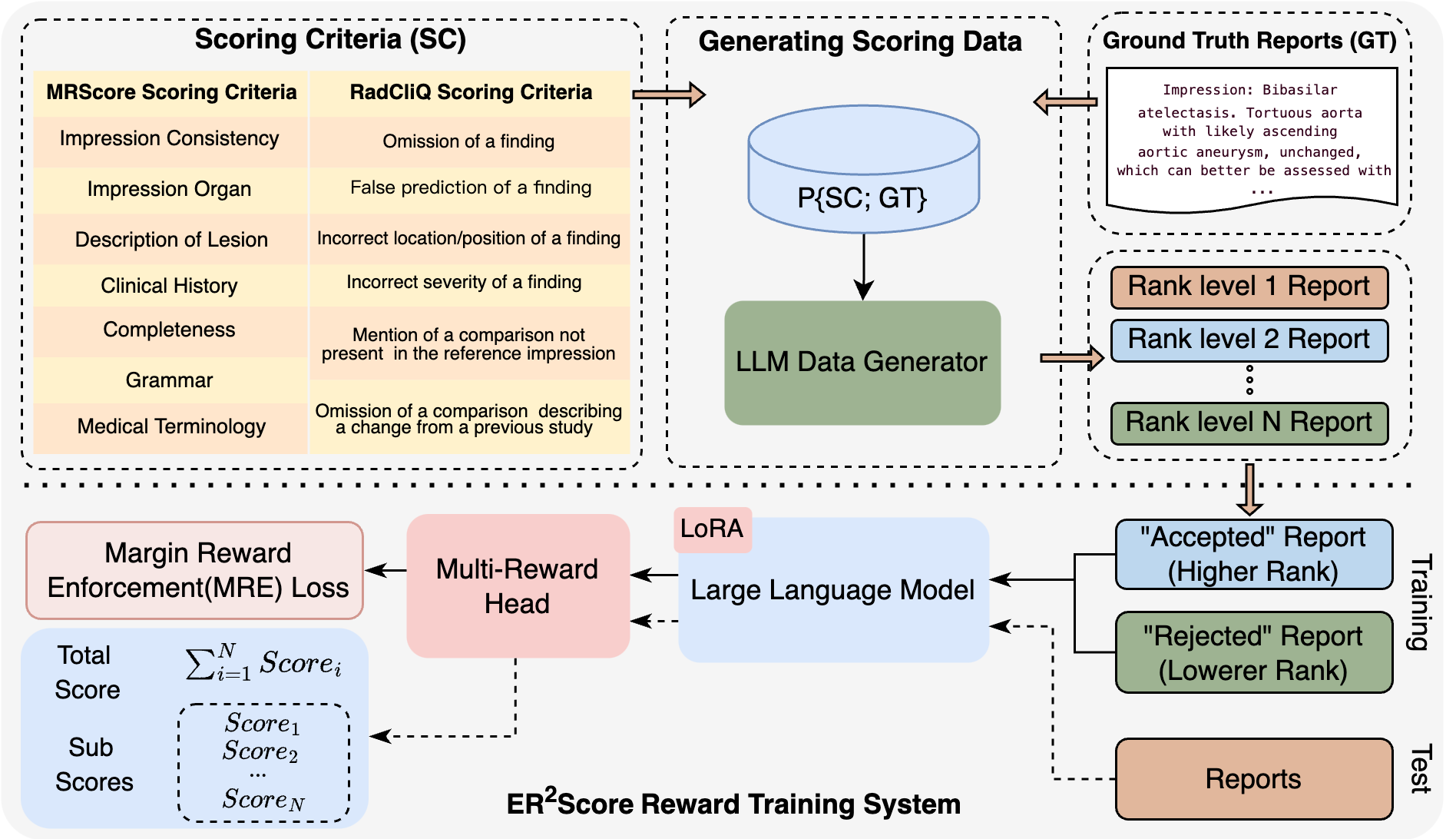}}
\caption{Overview of ReFINE. The upper portion illustrates the training data generation process, while the lower portion represents the training process for the reward model using LoRA. In the lower portion, the solid line indicates the training phase while the dashed line indicates the inference phase.}
\label{fig:MRScoreModel}
\end{figure*}

Our model not only provides an overall score but also delivers nuanced sub-scores to facilitate a more detailed interpretation of the assessment. This approach leverages GPT-4 to generate training samples by scoring $x$ against its reference $\hat{x}$ based on specified criteria. These samples are then paired up by our pairing rule and used to train a reward model with our proposed reward loss function to predict sub-scores. The summation of these sub-scores results in the final overall score. The overview of our framework is presented in Figure~\ref{fig:MRScoreModel}.

\subsection{Generating Training Data by GPT-4}\label{subsec:ScoringCriteria}
Recent studies have demonstrated GPT-4's capability in evaluating chest X-ray reports. When prompted with specified criteria, \textbf{GPT-4 can generate similarity assessments that statistically correlate with human evaluations}, as consistently verified in~\cite{chiang2023can} and~\cite{liu2024mrscore}. For example, in~\cite{chiang2023can}, GPT-4 achieved Kendall’s Tau of 0.735 with radiologists’ annotations using RadCliQ scoring system. In~\cite{liu2024mrscore}, GPT-4 scored a Kendall's Tau correlation of 0.531 with human ratings using the MRScore scoring system. Building on this observation, we utilize GPT-4 to generate extensive scoring data, including both reports and the corresponding scores, for training purposes. The process is elaborated as follows.

% \noindent \textbf{Training Data Evaluation} 
% We verified the quality of our training data by randomly selecting 50 GPT-4 generated training samples and having them evaluated by an experienced radiologist. The accuracies (accuracy = Total number of score samples that match human ratings / Total number of score samples) are 0.9 for Impression, 0.98 for Impression Organ, 0.86 for Description of Lesion, 0.92 for Clinical History, 0.98 for Completeness, 1.0 for Grammar, and 1.0 for Medical Terminology. 

\noindent \textbf{Defining Scoring Criteria.}~~Various assessment criteria have been reported in the literature. In this study, we investigate two scoring systems to demonstrate our model's versatility across different evaluation rules. The RadCliQ scoring system proposed in~\cite{yu2023evaluatingradcliq} evaluates both clinically significant and insignificant errors across six error categories: 1) false prediction of a finding, 2) omission of a finding, 3) incorrect location or position of a finding, 4) incorrect severity of a finding, 5) mention of a comparison absent in the reference impression, and 6) omission of a comparison that notes a change from a previous study. The total score is the sum of the error counts, highlighting the importance of clinical findings. Differently, the MRScore scoring system proposed in~\cite{liu2024mrscore} addresses both clinical findings and linguistic concerns. It involves seven fundamental items from radiologists’ expertise and literature review: ``impression consistency", ``impression organs", ``description of lesions," ``clinical history", ``completeness", ``grammar", and ``medical terminology", with a detailed explanation. Each item corresponds to an error type with yes/no answers and is assigned a different weight (from \{30, 20, 20, 10, 10, 5, 5\} accordingly) to form individual item scores.  The total score is calculated as $\text{Total\_score} = 100 - \sum_{i=1}^{7} S_i \times W_i$, where $S_i$ is error score of the $i$-th item and $W_i$ is the corresponding weight. With these scoring rules, GPT-4 can be prompted to score reports following these criteria, as elaborated below. 

\noindent \textbf{Generating Scoring Training Dataset.}~~With a defined scoring system, we craft prompts that encapsulate the evaluation criteria, guiding GPT-4 to assess radiology reports similarly to human evaluators. An example of a prompt can be found in the supplementary material. Utilizing the GPT-4 API, we generate reports of varying quality based on a randomly selected subset of ground-truth reports from the MIMIC-CXR dataset. For RadCliQ scoring, we randomly select around 8000 ground-truth reports, each leading to three GPT-4-generated reports reflecting varied error levels, i.e., 0-2 errors, 3-4 errors, and 5-6 errors. Each generated report is assessed for the total number of errors as well as individual error scores. Similarly, for the MRScore scoring system, we randomly select 1800 ground-truth reports, each with three GPT-4-generated reports corresponding to three quality tiers (0-40, 40-70, and 70-100). Each report is evaluated for both total quality and individual item scores. We verified the quality of our training data by randomly selecting 50 GPT-4 generated training samples and having them evaluated by an experienced radiologist. The accuracies (accuracy = Total number of score samples that match human ratings / Total number of score samples) are 0.9 for Impression, 0.98 for Impression Organ, 0.86 for Description of Lesion, 0.92 for Clinical History, 0.98 for Completeness, 1.0 for Grammar, and 1.0 for Medical Terminology.

\subsection{LLM-based Reward Model}
ReFINE is our innovative evaluation metric designed to be versatile across various evaluation frameworks. This LLM-based reward model leverages a pretrained language model, such as Llama3~\cite{touvron2023llama}, fine-tuning it to align with human evaluations using pairs of reports guided by our novel reward system. The core of ReFINE is its training process, which involves pairs of reports generated from the same ground-truth report but with different qualities. This pairing mechanism is essential for calibrating the model to distinguish between different quality levels effectively. During training, the model learns to assign higher rewards to high-quality reports while simultaneously generating multiple individual criterion scores. These criterion scores are critical as they provide detailed insights into specific aspects of the report's quality. At the inference stage, the model predicts rewards for each individual criterion. These rewards are then summed to generate the final ReFINE. To ensure precise differentiation, we also introduce a scoring margin for each criterion and the overall score. This margin enables the model to recognize and learn subtle differences in report quality, enhancing its evaluative capability.

\begin{figure*}[h!]
\centering
\centerline{\includegraphics[width=\linewidth]{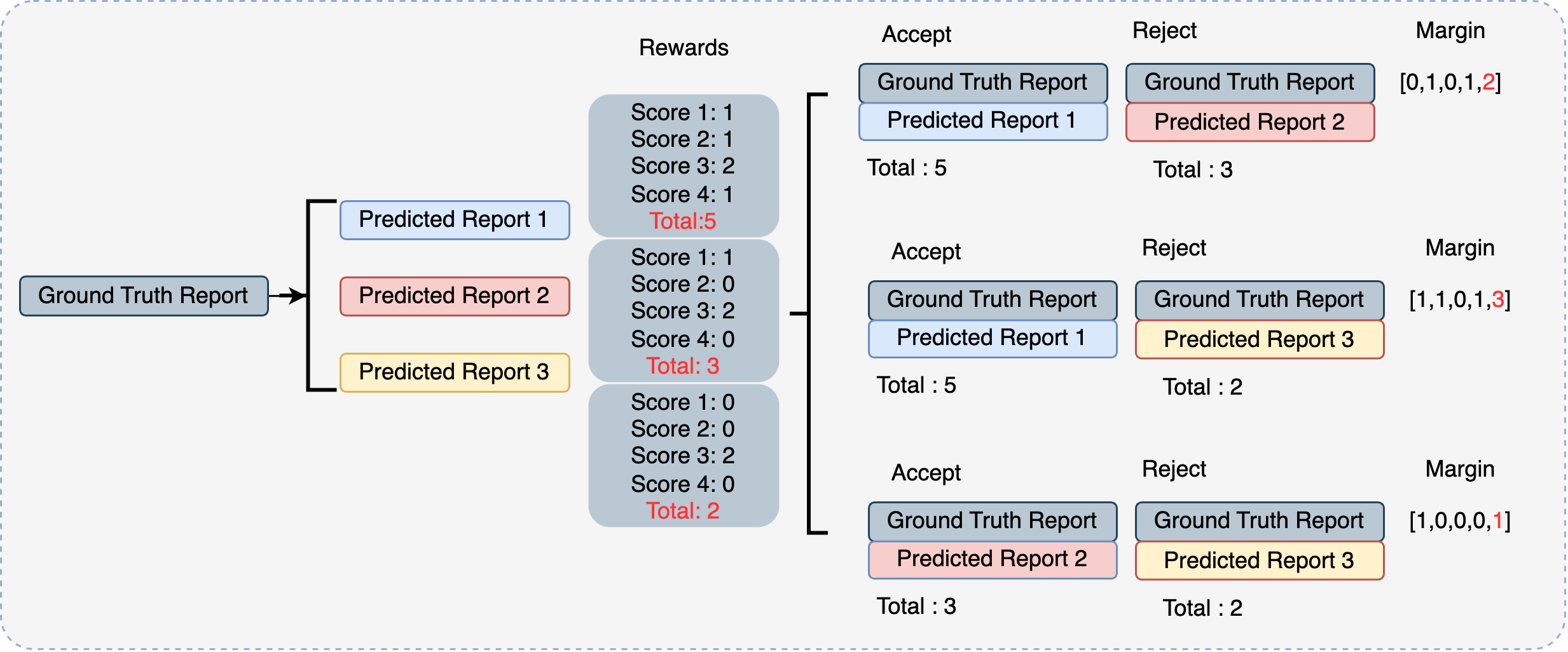}}
\caption{An illustration of report pairing rule, taking a scoring system with 4 criteria as an example.
}
\label{fig:DataPair}
\end{figure*}

\noindent \underline{\textbf {Model Input.}}~~Our model requires paired reports and their score margins as input. Each pair consists of an ``accepted" report and a ``rejected" report, both derived from the same ground-truth report, with the ``accepted" report having a higher score than the ``rejected" one. Figure \ref{fig:DataPair} illustrates the pairing rule, showing the selection process for accepted and rejected reports and the calculation of their respective margins.
In the example shown in Figure \ref{fig:DataPair}, a scoring system with four individual evaluation items is used. Accepted and rejected reports are determined based on their total scores. These reports, along with their ground-truth report, are then incorporated into a text prompt to fine-tune the LLM model for report assessment. In addition to the reports, we calculate a list of margins for both the four sub-scores and the total score: $margin^{i} = score_{accept}^{i} - score_{reject}^{i}$, where $i=1, \cdots, 5$ with $i=5$ corresponding to the total score and $i=1, \cdots, 4$ for sub-scores. A larger margin indicates a more pronounced quality discrepancy between the two reports, while a smaller margin suggests a lesser difference. Note that although the margin of the total score is always greater than 0, the margins of the sub-scores are not necessarily positive. 

\noindent \underline{\textbf{LLM Model.}}~~Our reward model, based on the Llama3~\cite{meta2024llama} backbone, incorporates a multi-reward head to generate the ReFINE. Llama3 was selected for its exceptional language comprehension with just 6.8M trainable parameters over 7 billion in total. The multi-reward head is a linear projection layer mapping Llama-3's last layer feature map to an $N\times1$ vector, where $N$ is the total number of sub-scores. This model is fine-tuned using Low-Rank Adaptation (LoRA)~\cite{hu2022lora} for parameter-efficient fine-tuning (PEFT), allowing effective fine-tuning with minimal parameter changes. Training pairs of "accepted" and "rejected" reports calibrate the model for reward prediction. During training, the model learns to distinguish report qualities by adhering to a scoring margin reflecting quality differences. Sub-scores discern quality differences per report aspect, with their summation producing the overall assessment for generated reports.

\noindent \underline{\textbf{Objective.}}~~Our multi-reward model aims to mimic human judgement via GPT-4 by optimizing a function based on the GPT-4 rankings of radiology reports. It discerns and predicts the preferred report within each pair, capturing subtle differences that distinguish superior reports.  Instead of rewarding based merely on the whole report, our objective function is devised to learn also the preference per individual criterion. The objective function is elaborated in Section~\ref{subsec:LossFunction}. Through our objective function, we can effectively utilize the total margin to control the overall quality of the report and also respect each sub-score's margin to manage the differences in sub-scores across varied overall quality levels. By adjusting the size of the margin, corresponding penalties are applied, thus training the model to produce appropriate rewards.
%\noindent \underline{\textbf {Fine-Tuning}}~~~During fine-tuning, the pre-trained LLM backbone model is tailored to suit the nuances of reward prediction, ensuring its outputs align with anticipated human evaluations. In our ReFINE training, we used LoRA\cite{hu2022lora} to fine-tune the Mistral-7B-instruct model by our training pairs, instructing it to differentiate between higher- and lower-quality reports while respecting the score margin between them: $margin = score_{accpet} - score_{reject}$. A larger margin indicates a more pronounced quality discrepancy between the two reports, while a smaller margin suggests a lesser difference.
%On top of the original Mistral model, we incorporated a reward head—a linear projection layer—that maps the LLM output features to a vector with the size of $N\times 1$, corresponding to the N individual criterion scores. The output reward scores are summed and utilized to evaluate the quality of the generated reports, serving as our ReFINE. 
%Subsequently, a sigmoid activation function is applied to scale each output between 0 and 1, where 1 signifies the highest reward and 0 represents the lowest. 
%The whole process is illustrated in the lower portion of Fig.~\ref{fig:MRScoreModel}.

\subsection{Margin Reward Enforcement (MRE) Loss Function}\label{subsec:LossFunction}
Considering a pair of generated reports $<y_w^i, y_l^i>$ \footnote{Here ``w" stands for ``win", indicating the accepted report, and ``l" for ``lose", indicating the rejected report.} corresponding to the same $i$-th ground truth report $x^i$, the accepted report $y_w^i$  receives a higher GPT-4  score $s_w^i$ and the rejected report $y_l^i$ a lower GPT-4 score $s_l^i$. Let $s_w^{i,j}$ and $s_l^{i,j}$ denote the $j$-th  sub-score of $s_w^i$ and $s_l^i$, respectively, where $j=1, \cdots, N$ and $N$ is the number of sub-scores for a specific scoring system. Note that although the total score $s_w^i$ is greater than $s_l^i$, the sub-score $s_w^{i,j}$ is not necessarily greater than $s_l^{i,j}$. Our objective is to train the model to discern the rankings of both individual and total scores of the report pair, formulated as follows:

\begin{align}
&{\mathcal L}_{ind} (y_w^i, y_l^i) = \frac{1}{N}\sum_{j=1}^{N}\mathbbm{1}(s_w^{i,j} \neq s_l^{i,j})ReLU(-t_w(r_w^{i,j}-r_l^{i,j})\nonumber\\ 
&  + t_wm^{i,j}) + (1-\mathbbm{1}(s_w^{i,j} \neq s_l^{i,j}))ReLU(|r_w^{i,j}-r_l^{i,j}| - c), \nonumber\\
&{\mathcal L}_{tot} (y_w^i, y_l^i) = ReLU(-(\sum_{j=1}^{N}r_w^{i,j} - \sum_{j=1}^{N}r_l^{i,j})+m^i),\nonumber\\
&{\mathcal L}_{MRE} = \sum_{i=1}^{K} {\mathcal L}_{ind} (y_w^i, y_l^i) + \lambda {\mathcal L}_{tot}(y_w^i, y_l^i).
\end{align}

Here $r_w^{i,j}$ and $r_l^{i,j}$ denote the $j$-th individual rewards assigned to the reports $y_w^i$ and $y_l^i$, respectively. The margin between the total scores $s_w^i$ and $s_l^i$ is denoted by $m^i = s_w^i - s_l^i$, where $m^i>0$. The individual ``margin" $m^{i,j} = s_w^{i,j}-s_l^{i,j}$ is not necessarily positive. The variable $t_w$ acts as a flag: $t_w = 1$ if $m^{i,j}>0$, otherwise $t_w = -1$. The function $\mathbbm{1} (\cdot)$ is an indicator function, returning 1 when the event occurs and 0 otherwise. $K$ is the total number of report pairs. 

Our overall loss ${\mathcal L}_{overall}$ comprises two terms: the individual reward loss ${\mathcal L}_{ind}$ and the total reward loss ${\mathcal L}_{tot}$, balanced by the hyperparameter $\lambda$. An analysis of the model's behavior is as follows. For the individual reward loss ${\mathcal L}_{ind}$, if the ground truth scores have the relationship of $s_w^{i,j} > s_l^{i,j}$, i.e., $m^{i,j}>0$, a penalty is incurred when the reward $r_l^{i,j}$ is larger than $r_w^{i,j}-m^{i,j}$; if $s_w^{i,j} < s_l^{i,j}$, i.e., $m^{i,j}<0$, a penalty is incurred when the reward $r_l^{i,j}$ is smaller than $r_w^{i,j}-m^{i,j}$; if $s_w^{i,j} = s_l^{i,j}$, a penalty is incurred when the absolute difference between the two rewards is larger than a preset small positive value $c$. In addition to minimizing the individual reward loss, we also regularize the total reward loss ${\mathcal L}_{tot}$, i.e., when the total reward $\sum_{j}r_l^{i,j}$ of the rejected report $y_l^i$ is larger than $\sum_{j}r_w^{i,j} - m^i$, a penalty is incurred. Minimizing ${\mathcal L}_{overall}$ ensures that our model furnishes both individual and total scores, thereby offering nuanced insights into the assessment results.

%===================================%
\section{Experiments and Result}
\subsection{Datasets}
We evaluated the effectiveness of ReFINE by assessing its alignment with expert radiologist evaluations, ensuring that its predictions correlate closely with those of human experts. Our evaluation involved two datasets, ReXVal~\cite{yu2023radiology} and Rad-100, each based on a distinct scoring system as described in Section \ref{subsec:ScoringCriteria}. This approach allowed us to validate ReFINE across different evaluative standards, exhibiting the model’s adaptability to diverse assessment systems.

\textbf{ReXVal} Dataset is a publicly accessible dataset that features six board-certified radiologists' evaluations of automatically generated radiology reports. It provides a comprehensive breakdown of clinically significant and insignificant errors across six distinct categories relative to the ground-truth reports drawn from the MIMIC-CXR dataset, i.e., the RadCliQ scoring system named in our paper. The dataset encompasses 200 pairs of candidate and ground-truth reports, derived from 50 studies, each generating four candidate reports. ReXVal is primarily utilized to assess the correlation between automated metric scores and human radiologist judgments, explore the limitations of current automated metrics, and develop an integrated metric for evaluating radiological report generation. 

\textbf{Rad-100} Dataset which we developed using the MRScore scoring system, consists of 100 diagnostic reports generated by the conventional R2Gen models. Each report displays varying qualities when compared to its corresponding ground-truth report, which has been randomly sampled from the MIMIC-CXR dataset. Employing this scoring system, an experienced radiologist performs detailed evaluations of each report, assessing both overall performance and individual criteria. These evaluations provide a robust foundation for validating our ReFINE.\footnote{The Rad-100 dataset is entirely distinct from the datasets used for training our reward model.}

\subsection{Performance on ReXVal Dataset}

\noindent \textbf{Correlation Analysis of Sub-criteria.}~~Table~\ref{tab:radcliq} provides a quantitative evaluation of ReFINE on the ReXVal dataset, specifically constructed based on the RadCliQ Scoring System. This assessment highlights significant alignment between ReFINE evaluations and expert radiologist judgments across various error categories, using Kendall’s Tau and Spearman Correlation coefficients as metrics. Notably, the high correlation scores in categories such as “False prediction of a finding” (Kendall’s Tau: 0.680, Spearman: 0.842) and “Omission of a finding” (Kendall’s Tau: 0.507, Spearman: 0.673) demonstrate ReFINE’s capability in accurately identifying common radiological errors, indicating its effectiveness in recognizing significant or typical lesions. Although ReFINE demonstrates strong correlations across most sub-criteria, there are areas for improvement. For example, the scores for “Incorrect location or position of a finding” (Kendall’s Tau: 0.246, Spearman: 0.327) are relatively low, possibly because location and position details are often subtle and challenging to capture accurately. It is worth noting that this also highlights the advantage of ReFINE over methods that provide only an overall score~\cite{yu2023evaluatingradcliq,zhang2019bertscore,jain2021radgraph}. By providing scores for each sub-criterion, ReFINE allows us to clearly identify specific areas where the model can be enhanced. The statistical significance of the results is underscored by extremely low p-values across all categories, reinforcing the robustness of the correlation between ReFINE and expert evaluations. The overall high scores—0.751 for Kendall’s Tau and 0.910 for Spearman Correlation—further validate the reliability of ReFINE as an evaluation tool, highlighting its potential utility in clinical and research settings for assessing radiology reports.

\begin{table}[h!]
\centering
\caption{Human Correlations of ReFINE on ReXVal Dataset using RadCliQ scoring criteria.}
\resizebox{\linewidth}{!}{
\begin{tabular}{@{}p{6cm}cc@{}}
\toprule
{\small \textbf{Criteria}} & {\small \textbf{Kendall’s Tau$\uparrow$(P-Value$\downarrow$)}} & {\small \textbf{Spearman$\uparrow$(P-Value$\downarrow$)}} \\ \midrule
- False prediction of a finding & 0.680 (9.0e-41) & 0.842 (6.2e-55) \\
- Omission of a finding & 0.507 (4.9e-23) & 0.673 (8.8e-28) \\
- Incorrect location or position of a finding & 0.246 (5.9e-6) & 0.327 (2.4e-6) \\
- Incorrect severity of a finding & 0.443 (4.6e-16) & 0.569 (1.5e-18) \\
- Mention of a comparison absent in the reference impression & 0.433 (4.6e-15) & 0.545 (7.3e-17) \\
- Omission of a comparison that notes a change from a previous study & 0.267 (1.4e-6) & 0.345 (5.7e-07) \\ \hline
Total &  0.751 (4e-52) & 0.910 (5e-76) \\
\bottomrule
\end{tabular}
}

\label{tab:radcliq}
\end{table}

\noindent \textbf{Comparison with other metrics.}~~Table~\ref{tab:metrics_comparison} compares the performance of different metrics using Kendall’s Tau and Spearman correlation on ReXVal Dataset. The comparison is based on the total score. Unlike ReFINE, \textit{the existing metrics cannot be customized to user-specific sub-criteria}, making sub-score comparison impossible\footnote{GREEN only provides error counts for each subcategory without human correlation, making direct comparison of subscore correlations unfeasible. Additionally, its overall correlation with human assessments is significantly lower than ours.}.

We evaluate our ReFINE against various NLG metrics, including BLEU-4~\cite{papineni2002bleu}, ROUGE-L~\cite{lin2004rouge}, METEOR~\cite{banerjee2005meteor}, and CIDEr~\cite{vedantam2015cider}, as well as clinical metrics like BERTScore~\cite{zhang2019bertscore} and RadGraph F1~\cite{jain2021radgraph}. We also compare with RadCliQ-based metrics~\cite{yu2023evaluatingradcliq} derived from human-annotated error scores, as well as the LLM-based GREEN score.

The table demonstrates that ReFINE exhibits a strong alignment with human judgments, as evidenced by its Kendall’s Tau value of 0.751 and Spearman correlation of 0.910, both surpassing all other evaluated metrics. For instance, traditional NLG metrics like BLEU-4, ROUGE-L, and METEOR show lower correlations, with BLEU-4 achieving a Kendall’s Tau of 0.345 and a Spearman correlation of 0.475. Similarly, clinical metrics such as BERTScore and RadGraph F1, while performing better than traditional NLG metrics, still fall short compared to ReFINE. %BERTScore, for example, has a Kendall’s Tau of 0.507 and a Spearman correlation of 0.677. 
RadCliQ-v1 metric shows high correlation values, with a Kendall’s Tau of 0.631 and a Spearman correlation of 0.816, indicating its effectiveness in aligning with human evaluations. The GREEN score shows higher correlation values, with Kendall’s Tau 0.640, lower than our 0.751. GREEN requires 1.06 seconds to infer one sample, while our model only needs 0.04 seconds. Notably, GREEN incurs much higher training cost by using 8 NVIDIA A100 GPUs (40GB VRAM) with a batch size of 2,048 for 12 epochs, whereas our model was trained with only 1 A6000 GPU (40GB VRAM) with a batch size of 6 for 4 epochs. However, our ReFINE outperforms all these metrics, highlighting its superior ability to capture the nuances of radiology report generation as judged by experts. 

% ER²Score
\begin{table}[h!]
\centering
\caption{Human Correlation Comparison of Evaluation Metrics on ReXVal Dataset}
\resizebox{0.8\linewidth}{!}{
\begin{tabular}{@{}lcc@{}}
\toprule
{\small \textbf{Metric}} & {\small \textbf{Kendall’s Tau$\uparrow$(P-Value$\downarrow$)}}  & {\small \textbf{Spearman$\uparrow$ (P-Value$\downarrow$)}} \\
\midrule
BLEU-4~\cite{papineni2002bleu} & 0.345 (2.2e-12) & 0.475 (1.2e-12) \\
ROUGE-L~\cite{lin2004rouge} & 0.491 (2.9e-23) & 0.663 (1.2e-26) \\
METEOR~\cite{banerjee2005meteor} & 0.464 (8.4e-21) & 0.627 (2.8e-23) \\
CIDEr~\cite{vedantam2015cider} & 0.499 (4.5e-24) & 0.664 (8.9e-27) \\
BertScore~\cite{zhang2019bertscore} & 0.507 (4.5e-25) & 0.677 (3.9e-28) \\
RadGraphF1~\cite{jain2021radgraph} & 0.516 (4.3e-25) & 0.702 (4.4e-31) \\
semb\_score~\cite{yu2023evaluatingradcliq} & 0.494 (1.0e-23) & 0.665 (6.2e-27) \\
RadCliQ-v1~\cite{yu2023evaluatingradcliq} & 0.631 (6.9e-38) & 0.816 (6.6e-49) \\ 
GREEN~\cite{ostmeier2024green} & 0.640  & - \\ \hline
ReFINE (Ours)  & \textbf{0.751 (4.0e-52)} & \textbf{0.910 (5.0e-76)} \\
\bottomrule
\end{tabular}
}
\label{tab:metrics_comparison}
\end{table}

%======================================================%
\subsection{Performance on Rad-100 Dataest}

\noindent \textbf{Accuracy analysis of sub-criteria.}~~Since the scoring system used by Rad-100 is a binary format where the presence of an error is marked as 1 and the absence as 0 (check supplementary for detail), the results are multiplied by pre-defined weights before forming the final score. Accordingly, we evaluate the accuracy of binary classification for each sub-criterion, as reported in Table~\ref{tab:sub_scores_accuracy}.

% \begin{table}[h!]
% \centering
% \caption{Accuracy of Different Sub-scores in GPT-Rad test dataset}
% \begin{tabular}{@{}lc@{}}
% \toprule
% \textbf{Sub-score} & \textbf{Accuracy} \\
% \midrule
% Impression Consistency & 0.589 \\
% Impression Organ & 0.730 \\
% Description of Lesion & 0.770 \\
% Clinical History & 0.410 \\
% Completeness & 0.380 \\
% Grammar & 0.980 \\
% Medical Terminology & 0.720 \\
% \bottomrule
% \end{tabular}
% \label{tab:sub_scores_accuracy}
% \end{table}

\newcommand{\smallbf}[1]{\textbf{\small #1}}
\begin{table*}[h!]
\centering
\caption{Accuracy of Different Sub-scores in Rad-100 test dataset. Here, `Imp. Cons.' stands for Impression Consistency, `Imp. Org.' for Impression Organ, `Desc. Les.' for Description of Lesion, `Clin. Hist.' for Clinical History, `Comp.' for Completeness, `Gram.' for Grammar, and `Med. Term.' for Medical Terminology.}
\begin{tabular}{@{}lccccccc@{}}
\toprule
\smallbf{Sub-criteria} & \smallbf{Imp. Cons.} & \smallbf{Imp. Org.} & \smallbf{Desc. Les.} & \smallbf{Clin. Hist.} & \smallbf{Comp.} & \smallbf{Gram.} & \smallbf{Med. Term.} \\
\midrule
\textbf{Accuracy} & 0.589 & 0.730 & 0.770 & 0.410 & 0.380 & 0.980 & 0.720 \\
\bottomrule
\end{tabular}
\label{tab:sub_scores_accuracy}
\end{table*}

\noindent \textbf{Comparison with other metrics.}~~Table~\ref{tab:metrics_comparation_gpt4} provides a performance comparison of metrics using Kendall’s Tau and Spearman correlation on the Rad-100 dataset. Similar to the previous analysis on the ReXVal dataset, we evaluate our ER²Score against various NLG and clinical metrics. As observed, on the Rad-100 dataset, our ER²Score demonstrates superior performance, with a Kendall’s Tau of 0.230 and a Spearman correlation of 0.293, both statistically significant with a p-value of 0.003.

% The table~\ref{tab:metrics_comparation_gpt4} illustrates the performance of various metrics in terms of their Spearman and Kendall Tau correlations with human judgment scores. 
% ReFINE stands out significantly in terms of its Spearman and Kendall Tau correlations. It shows the highest correlation values among all metrics, with a Spearman correlation of 
% 0.293
% 0.293 and a Kendall Tau correlation of 
% 0.230
% 0.230. These are statistically significant with p-values (
% 0.003
% 0.003), indicating strong reliability and alignment with human judgments.

\begin{table}[h!]
\centering
\caption{Human Correlation Comparison of Evaluation Metrics on Rad-100 Dataset}
\resizebox{0.8\linewidth}{!}{
\begin{tabular}{@{}lcc@{}}
\toprule
{\small \textbf{Metric}} & {\small \textbf{Kendall’s Tau$\uparrow$(P-Value$\downarrow$)}}  & {\small \textbf{Spearman$\uparrow$(P-Value$\downarrow$)}} \\
\midrule
BLEU-4~\cite{papineni2002bleu} & 0.07 (0.49) & 0.05 (0.51) \\
ROUGE-L~\cite{lin2004rouge} & 0.16 (0.10) & 0.12 (0.10) \\
METEOR~\cite{banerjee2005meteor} & 0.11 (0.27) & 0.08 (0.26) \\
CIDEr~\cite{vedantam2015cider} & 0.04 (0.70) & 0.03 (0.65) \\
BertScore~\cite{zhang2019bertscore} & 0.13 (0.19) & 0.09 (0.20) \\
RadGraphF1~\cite{jain2021radgraph} & 0.09 (0.38) & 0.06 (0.43) \\
semb\_score~\cite{yu2023evaluatingradcliq} & 0.01 (0.94) & 0.01(0.94) \\
RadCliQ-v1~\cite{yu2023evaluatingradcliq} & 0.08(0.44) & 0.06 (0.45) \\ \hline
Ours(ReFINE)  & \textbf{0.23 (0.003)} & \textbf{0.29 (0.003)} \\
\bottomrule
\end{tabular}
}
\label{tab:metrics_comparation_gpt4}
\end{table}

\subsection{Performance Comparison of LLM backbones}
Table~\ref{tab:model_comparison} presents a performance comparison of various LLM backbones. Notably, Llama3 demonstrates superior performance with a medium size of trainable parameters. To ensure the scoring system is easily deployable, we focused on models with 7 billion parameters in total or fewer.
\begin{table}[h!]
\centering
\caption{Ablation Study of LLM Backbones on ReXVal Dataset}
\resizebox{0.8\textwidth}{!}{
\begin{tabular}{@{}l|ccc@{}}
\toprule
\textbf{Model} & \textbf{Trainable Params (\%)} & \textbf{Kendall's Tau (\(\uparrow\)))} & \textbf{Spearman (\(\uparrow\))} \\
\midrule
Llama3 ~\cite{meta2024llama}         & 6.8M (0.090) & 0.751  & 0.910  \\
Vicuna-7b~\cite{chiang2023vicuna}     & 8.4M (0.127) & 0.738 & 0.901 \\
Meditron ~\cite{chen2023meditron}       & 8.4M (0.127) & 0.709  & 0.880 \\
Gemma-7b ~\cite{gemma_2024}       & 6.4M (0.075) & 0.707  & 0.876 \\
% BioMistral     & 6.8M (0.096) & 0.699 (6.7e-46) & 0.867 (1.1e-61) \\
Qwen1.5-7b\cite{qwen}    & 8.4M (0.110) & 0.684  & 0.858 \\
Phi-2 ~\cite{li2023phi2}       & 5.3M (0.196) & 0.591  & 0.784  \\
\bottomrule
\end{tabular}
}
\label{tab:model_comparison}
\end{table}

\subsection{Ablation study of losses and hyperparameters}
The loss we proposed comprises two terms: the individual reward loss \( L_{\text{ind}} \) and the total reward loss \( L_{\text{tot}} \). An ablation of the loss functions is given in Table \ref{table:loss_ablation}. As shown, if we train  \( L_{\text{tot}} \) alone for predicting sub-scores, Kendall's Tau will drop from 0.751 to 0.740 for the total score, a sum of the sub-scores. If we train \( L_{\text{ind}} \) alone, Kendall's Tau will drop from 0.751 to 0.738, demonstrating the effectiveness of the regularization from \( L_{\text{tot}} \).

% \begin{table}[h!]
% \centering
% \caption{Spearman and Kendall correlation coefficients for different methodologies}
% \begin{tabular}{@{}lcc@{}}
% \toprule
% \textbf{Methodology} & \textbf{Spearman (\(\uparrow\))%(P-Value (\(\downarrow\))
% } 
% & \textbf{Kendall Tau (\(\uparrow\))%(P-Value (\(\downarrow\))
% }  \\
% \midrule
% \({\mathcal L}_{MRE}\)  (Ours complete) & 0.910  & 0.751  \\
% \({\mathcal L}_{tot} %(y_w^i, y_l^i)
% \)  Only & 0.899 & 0.740  \\
% \({\mathcal{L}}_{\text{ind}} %(y_w^i, y_l^i)
% \)  Only & 0.899 & 0.738  \\
% \bottomrule
% \end{tabular}

% \label{table:loss_ablation}
% \end{table}

\begin{table}[h!]
\centering
\caption{Spearman and Kendall correlation coefficients for different methodologies}
\resizebox{0.5\textwidth}{!}{
\begin{tabular}{@{}c|c|ccc@{}} % Horizontal lines at top, after header, and bottom with specific thickness
\hline
$\mathcal{L}_{\text{tot}}$ & $\mathcal{L}_{\text{ind}}$ & \textbf{Spearman (\(\uparrow\))} & \textbf{Kendall's Tau (\(\uparrow\))} \\ \hline
\checkmark &                & 0.899 & 0.740 \\
           & \checkmark     & 0.899 & 0.738 \\
\checkmark & \checkmark     & 0.910 & 0.751 \\ \hline
\end{tabular}
}
\label{table:loss_ablation}
\end{table}

% \begin{table}[h!]
% \centering
% \caption{Spearman and Kendall correlation coefficients for different methodologies with loss components}
% \begin{tabular}{@{}lcccc@{}}
% \toprule
% \textbf{Methodology} & \textbf{Spearman (\(\uparrow\))} & \textbf{Kendall Tau (\(\uparrow\))} & \(\mathcal{L}_{tot}\) & \(\mathcal{L}_{ind}\) \\
% \midrule
% \(\mathcal{L}_{MRE}\) (Ours complete) & 0.910 & 0.751 & \(\times\) & \(\times\) \\
% \(\mathcal{L}_{tot}\) Only & 0.899 & 0.740 & \(\times\) &  \\
% \(\mathcal{L}_{ind}\) Only & 0.899 & 0.738 &  & \(\times\) \\
% \bottomrule
% \end{tabular}
% \label{table:loss_ablation_with_x}
% \end{table}

Our loss function involves two hyper-parameters: the hyperparameter \( c \) is just a small positive rounding number when judging whether \( r_w \) equals \( r_l \), which we set to 1e-2. The hyperparameter \( \lambda \) balances the two loss terms \( L_{\text{ind}} \) and \( L_{\text{tot}} \) and we examined its effect through the ablation study shown in Table\ref{table:lambda_ablation}. As seen, our model is insensitive to \( \lambda \). When it varies in a reasonable range, our model produces better human correlations than the existing evaluation metrics.

\begin{table}[h!]
\centering
\caption{Spearman and Kendall correlation coefficients with varying \( \lambda \) values}
\resizebox{0.45\textwidth}{!}{
\begin{tabular}{@{}l|cccccc@{}}
\toprule
\( \lambda \) & 0.5 & 0.8 & 1.0 & 1.2 & 2.0 & 3.0 \\
\midrule
Spearman & 0.904 & 0.906 & 0.910 & 0.900 & 0.895 & 0.893 \\
Kendall & 0.743 & 0.746 & 0.751 & 0.740 & 0.735 & 0.729 \\
\bottomrule
\end{tabular}
}
\label{table:lambda_ablation}
\end{table}

\subsection{Qualitative Analysis}
\begin{figure}[h!]
\centering
\centerline{\includegraphics[width=\linewidth]{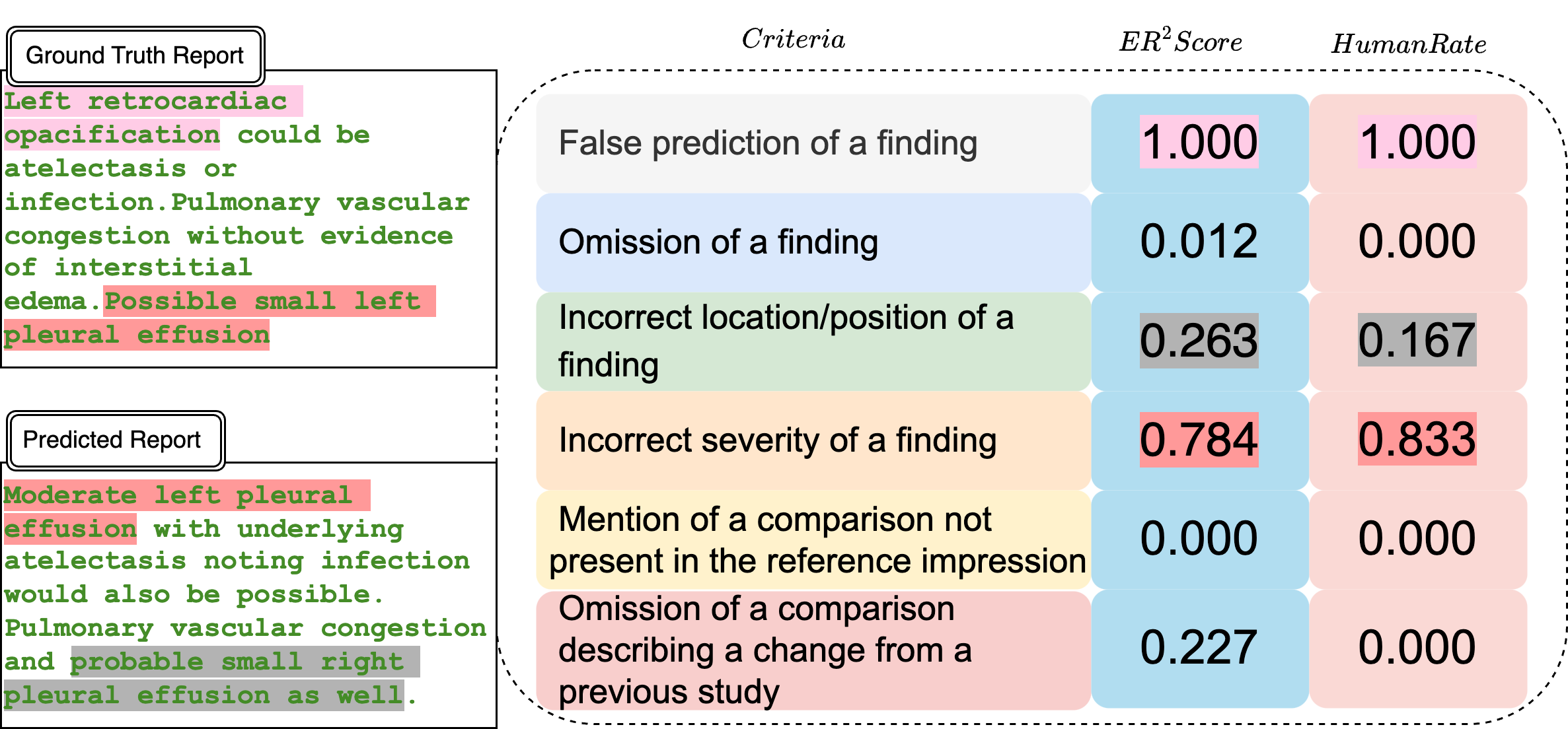}}
\caption{An visual example of ReFINE from ReXVal Dataset. The highlighted sentences in reports and their corresponding scores share the same colors.}
\label{fig:RadCase}
\end{figure}

A visual example is provided in Figure~\ref{fig:RadCase}, demonstrating how the ReFINE correlates with human ratings using the RadCliQ scoring system. As shown, the generated report inaccurately describes the severity of the ``left pleural effusion'' (highlighted in red), resulting in a high ReFINE for ``incorrect severity of a finding'', which aligns with the human rating. Additionally, the report erroneously mentions a ``right pleural effusion'', leading to an ``incorrect location/position of a finding'', again perceived similarly by both the ReFINE and human ratings. Lastly, the generated report fails to mention the ``left retrocardiac opacification'',  leading to a score of `1.0' for ``false prediction of a finding'' from both the ReFINE and the human rating.

\section{Conclusions}
\label{sec:conclusions}

ReFINE offers a human-correlated and explainable metric for evaluating radiology reports. It allows for more fine-grained scoring, aligning each item of the evaluation rule with its respective sub-score, therefore enhancing the interpretability of assessment results. Leveraging GPT-4’s human-like scoring capacity, we have tailored 
extensive training samples to fine-tune LLMs toward discerning report qualities using our designed reward loss. Our metric’s adaptability allows for accommodating various scoring criteria. 

Our method has the following 
\textbf{limitations}. First, the level of explainability could be improved by adding detailed paragraph explanations, which are not currently included. Second, due to the costly nature of human evaluation, the scale of the test sets in this study remains limited. However, we need to emphasize that the scale of datasets in this work matches that used in comparable works in the literature. 
%Third, while MIMIC-CXR is a comprehensive benchmark for chest X-rays, potential biases in the dataset could affect our model, warranting further exploration in future work. 

%============================%
% \bibliographystyle{plainnat}
% \bibliographystyle{unsrt}
% \newpage
% \bibliography{refs}

\section{Ethics Statement}
 Our ReFINE model, which fine-tunes LLAMA-3 as a reward system, operates entirely locally once trained, eliminating the need for any interactions with GPT-4 during inference. This local deployment ensures that there is no risk of information leakage. GPT-4 is only used to generate training data from MIMIC-CXR dataset. MIMIC-CXR is a public dataset, which has been anonymized and de-identified. The platform Azure OpenAI is HIPAA compliant and ensures the privacy and compliance of medical data (e.g., the data are not accessible to OpenAI).
 
\newpage
\bibliographystyle{unsrt}  
% \bibliography{main}
\bibliography{main}
\end{document}